# AGENT MODELS OF POLITICAL INTERACTIONS
Eric Engle

Eric Engle

# INTRODUCTION

Emergent qualities of a system of actors (agents) are those resultants from the interactions of those agents.[1] These group interactions can appear to an outside observer to mimic the intelligence of a single entity.

In social sciences some prominent theorists have discussed emergence - perhaps without even realising it. In all events, many social phenomena can be modeled using emergence. This paper discusses emergence as a tool of political analysis, examines existing political simulations, and proposes a simple simulation using an agent model to determine emergent characteristics of a political system.

The paper is divided into three sections: First, a description of emergence in social science. Second, a description of relevant computer science concepts, including existing implementations. And finally, a brief description of the author's implementation.

# I. SOCIAL SCIENCE
## A. Emergence in Social Sciences

Adam Smith described the free market as an emergent phenomenon: each individual trader seeking to maximizes his own well being acts like an agent with a utility function. The economic agent, acting rationally to maximize their utility function, negotiates transactions with other agents who also have utility functions. The agents exchange reward and punishment messages through transactions, and seek in each transaction to maximize self interest, represented through personal wealth. Smith observed that out of these self-interested transactions emerged, according to Smith, the optimal distribution of wealth.[2]

---

[1] For a definition of emergence see: Wikipedia "Emergence" (2004) http://en.wikipedia.org/wiki/Emergence

Smith also noted the specialization of his economic agents[3]: each agent has a specific economic function. Specialization allows each agent to concentrate its limited resources on solving one part of a larger problem, relying on the other agents to solve remaining parts of the problem. Specialized functions of individual agents explain why a series of individual transactions results in greater social wealth than the isolated acts of individuals. An agent focused on one activity becomes much better at it then another agent who focuses on many activities.

Ricardo takes the analysis of Smith even further, pointing out that increased productivity through specialisation arises not only through individual transactions but also as a result of whole nations trading peacefully with each other.[4] Ricardo demonstrates that even where one agent is absolutely inferior to another it is still to the profit of both agents to trade because relatively the absolutely inferior agent will be less inferior in some specialized economic function.

Emergence also can be seen in political science. According to Hobbes, the state arises through consent of persons living in a stateless condition and subject only to the law of the jungle, the law of the strongest. Hobbe's individual political agent surrenders individual power *via* a voluntary transaction with other individuals who also renounce their individual power to form the state. Hobbes' state is composed voluntarily by the transactions of persons living in it. Out of these individual transactions of real persons an artificial person with an artificial soul known as 'sovereignty' comes into being.[5] Hobbes is also describing emergence: in Hobbes' state each individual is an agent, entering into transactions to contract with other individuals to form the state.

---

[2] See, Adam Smith, *On the Nature and Causes of the Wealth of Nations* (1776) http://www.econlib.org/library/Smith/smWN.html
[3] *Id.* Book I, Chapter I note 39.
[4] David Ricardo, *On The Principles of Political Economy and Taxation*, Ch. 7 (1817) http://www.marxists.org/reference/subject/economics/ricardo/tax/ch07.htm

A third example of emergence in social theory is the idea of 'the balance of power'.[6] According to the balance of power theory, states seek to form coalitions with other states to form a system to prevent the emergence of any one state as dominant hegemon. This is exactly a problem of agent theory in artificial intelligence: when and how do competing agents enter into coalitions to achieve their individual goals? According to the proponents of the balance of power theory, the constant shifting alliances of states prevent the emergence of a dominant tyrannical power and maintain peace thereby. This description is another example of an emergent phenomenon.

But if social science does describe emergence, the emergence described in theory is sometimes not proven in practice. The balance of power theory is a good example of this. The theory was well worked out[7] and seemed plausible, but ultimately did not correspond to reality. Rather than preserve peace through uncertainty and prevent the emergence of one dominant power the shifting alliances of the international system resulted in two world wars. Further, the result was the ultimate domination of the world not by a shifting coalition of states but rather by two, then one hegemonic powers. These facts invalidate the balance of power theory.

---

[5] Thomas Hobbes, Leviathan, ch. 14 (1660) http://oregonstate.edu/instruct/phl302/texts/hobbes/leviathan-c.html#CHAPTERXIV

[6] Hume, David *Essays, Moral, Political, and Literary* Part II, Essay VII, "Of the Balance of Power" (1742) http://www.econlib.org/library/LFBooks/Hume/hmMPL30.html#Part%20II,%20Essay%20VII,%20OF%20THE%20BALANCE%20OF%20POWER

[7] See, Wikipedia "Balance of Power" (2004) http://en.wikipedia.org/wiki/Balance_of_power

The social contract theory can also be criticized as not corresponding to reality. No social contract or state of nature ever existed because in fact the state emerges from tribes, which emerged from extended families, which emerged from families. However, this Aristotelian conception of the rise of the state is still an example of emergence since the different groups involved have collective interests which they seek to advance and defend by entering into negotiations and transactions with other groups.

The best prominent example of emergence in social sciences appears to be the economics of Adam Smith. Economics can be clearly analyzed, as Smith does, as an emergent phenomenon. In contrast, emergence theory in international relations has not recovered from the failure of balance of power theory to describe and predict reality. This leads to the interesting observation that emergence theory in the human sciences has been most accurate in the positive sum game of economic relations and least successful in the zero sum game of deterrence and the negative sum "game" of war.

This paper intends to develop a simple model of international relations as an emergent phenomenon. First, we must see that the military aspects of international relations are, aside from humanitarian relief zero sum or even negative sum, while the economic aspects of international relations are positive sum. At the same time we must recognize that states wish to maximize both their power (military strength) and their wealth (economic well being). We must also recognize that some acts which maximize power minimize economic well being and vice verse. Further, we must describe and develop competing strategies. Clearly some states are mercantile states: historically and even to a lesser extent today Singapore, Bremen, Hamburg, the Netherlands and Venice clearly centered their foreign policy on trade. Other states have centered their strategy on military security: Sparta, Israel, Prussia and the U.S.S.R. are examples of national security states. Other states such as the United States Britain, have relied on flexible policies of commerce and conquest to advance their interests.

Are the emergent qualities of the international system healthy or dysfunctional? Yes: Emergence in the economy is clearly healthy, but emergent qualities of the international system in power dynamics such as arms races and war are clearly dysfunctional.

**B. The contemporary international system**

The contemporary international system is hegemonic: one power, the United States, dominates the globe, at least at present, both economically and militarily. At the same time however, the concept of sovereignty is breaking down. Sovereignty is simultaneously a) devolving through privatization, regionialism, and the rising importance of individual actors whether corporations or individuals and b) sublimating into transnational political entities such as the World Trade Organization, the European Union, MERCOSUR, the United Nations, and NATO - among others. Sovereignty has weakened due to globalisation.[8] Thus, the international system is increasingly complex.[9] This system appears to display emergent properties of dynamic stability: dynamic, in that economic restructuring happens consstantly. And stable, for precisely that reason. Economic dynamism obviates predatory wars. This complex system has been analysed using tools familiar to computer science: game theory, agency and emergence.

**II. COMPUTER SCIENCE**
**A. AI in Game Theory**
    **1. Game Theory**

---

[8] Andrew Grosso, *The Demise of Sovereignty*, 44/3 Communications of the ACM (2001) p. 102.
[9] "The main trend in the postwar international system is proliferating complexity in all dimensions of analysis and a parallel information explosion." John Mallery, "Thinking about Foreign Policy: Finding an Appropriate Role for Artificially Intelligent Computers", 1998 Annual Meeting of the International Studies Association (1988)

Games may be classified as zero sum (where a win for one player translates into a loss for other players) positive sum (where a gain for one player is a gain for other players) or negative sum (where each player is in absolute terms worse after the conclusion of the game). The central point of games is strategy: each player must choose a strategy from a set of possible strategies. Each chosen strategy will have some payoff. The central question of game theory is what constitutes rational behavior? Rationality is usually defined as that strategy which a player will adopt and benefit most from regardless of the strategies chosen by other players.[10]

## 2. Coalitions

Multiplayer games, such as international politics, create the opportunity for competing players to form temporary coalitions to secure advantages that they could not achieve alone. "A coalition is a set of self-interested agents that agree to cooperate to execute a task or achieve a goal"[11] An important question of coalition theory is the ability of the coalition to perform a task. If the coalition can perform the task better than all the individual agents alone then the agents have every reason to form a coalition. "A coalition's ability to perform a task can be determined by the abilities of its member agents to perform their respective subtasks"[12]

## 3. Coalitional Game Theory

---

[10] "In a game each of n players can choose among a set of strategies $S_i$, i=1 ... n, and there are functions $U_i$, i=1...n:$S_1$ x....x $S_n$ -->R which assign to each such combined choice a payoff for each player. The fundamental question of Game Theory is, what constitutes rational behavior in such a situation? The predominanct concept of rationality here (but by no means the only one) ius the Nash equilibrium: A combination of strategies $X_1$ E $S_1$,...$X_n$ E $S_n$ for which $U_1(X_1...X_i...X_n)$ >= $U_i(X_1...X'_i...X_n)$ for all i and P749 X' E $S_i$: a behavior, that is from which no player has an incentive to deviate."
Christos Papadimitriou, *Algorithms, Games and the Internet,* STOC'01 (2001), p. 749-750  2001 ACM 1-58113-349-9/01/0007.
[11] Ted Scully, Micheal Maden, Gerard Lyons "Coalition Calculation in a Dynamic Agent Environment" Proceedings of the 21st International Conference on Machine Learning, Banff, Canada (2004).
[12] *Id*.

The problem of game theory is determining what constitutes a rational strategy (e.g. MiniMax). The problem of coalitional game theory is to determine: First, whether to form a coalition (the payoff of all coalition members is better than that payoff any coalition member would have had on its own). Once it is known that the payoff of the coalition would be superior an important related question is what payoff the members of the coalition is optimum. These propositions can be mathematically represented as follows:

> "Coalitional game theory considers a game of n players as a set of possible $2^n$-1 coalitions, each of which, call it S, can achieve a particular value v(S) (the best possible sum of payoffs among players in S, against worst case behavior of players in [n]-S). The problem is how to divide the total payoff v([n]) among the n players. Many such notions of 'fairness' have been proposed, defended and criticized over the past decades. ...x, considered as a proposed splitting of the total payoff v([n]) among the n players is fair according to the core school if no coalition has an incentive to secede (because no coalition can make more by itself than it is allocated in x)."[13]

### 4. Opponent Modeling

While game theory can develop an ideal strategy against all opponents, opponent modeling seeks to develop the ideal strategy against this particular opponent. Opponent modelling is important in games where information is imperfect. Further, while an algorithm such as MiniMax will always perform optimally in a game with perfect information a faster strategy requiring fewer iterations may result in a greater payoff for the game. That is, if a game permits not only "win" or "loss" but also "score" opponent modeling might allow a faster win and/or earn a higher score than MiniMax alone.

---

[13] Christos Papadimitriou, *Algorithms, Games and the Internet,* STOC'01 (2001) p. 750. 2001 ACM 1-58113-349-9/01/0007

Opponent modeling is also important because it reveals our unstated assumptions regarding our opponents. An analysis of literature in game theory and strategies reveals a very disturbing fact: researchers tend to make the erroneous assumption that opponents will tend to act alike.[14] It is natural to assume that others will act as we do. In a symmetrical conflict this is also correct. But in asymmetrical conflicts presuming our opponent will act as we do is an error. Different players have differing goals and means of achieving them.

A similar error to presuming our opponent sees the world we do is failing to plan for conflict termination. The current war in Iraq is a perfect example of that recurring problem, however earlier wars such as the Suez War of 1956 or the Korean war and even the Japanese strategy in the Second World War also show the problem of failing to plan for war termination. People tend to assume the best and to assume that which confirms their beliefs: consequently, they fail to plan for events which do not cohere with their pre-existing notions. A related problem is modelling defeat. If states fail to plan for the end of the wars they start it may be because their models of defeat are inadequate.

Defeat is defined as that point where a combatant lacks sufficient emotional or physical resources to continue fighting. Modeling defeat reveals tactical asymmetries: sometimes a limited application of force can result in wide ranging effects.[15] This is because defeat by attrition, i.e. exhaustion, is not in fact the only model of defeat: sometimes combatants suddenly break down due to lack of support, communications, or bad leadership.[16] A consequent hypothesis is that states which fail to plan for war termination either expect to inflict a sudden non-attritional defeat or believe that only defeat by attrition is possible. In all events, there is a lack of models of psychological and political effects on conflict.[17] Material factors such as fatigue are more readily analysed but not necessarily more important. Fatigue occurs not just when a combatant is damaged but more importantly when it is damaged more severely than its opponent.[18] Failure to plan for war termination and the consequence of nearly equal levels of losses in attritional warfare partly explain why the first world war lasted so long.

---

[14] "On the understanding of Kennedy's problem representation they had coded into JFK/CUBA, the model rejected as implausible Dean Rusk's view that the soviets would likely retaliate with military action against West Berlin for any American airstrikes in Cuba. Because Rusk viewed the American dilemma regarding Cuba as essentially symmetrical to the Societ quandary regarding West Berlin, he could reasonably infer Soviet retaliation on West Berlin. Since the productions used to represent Kennedy's problem representation included no such symmetry, it ruled out Rusk's scenario." In fact the USSR and US were asymmetric opponents. The US was a status quo mixed power, relying as much on mercantile as on military strength. The USSR was a non-status quo military power. Further, they were in fact very unequal powers in terms of their disposable wealth and military capacity (the US had an absolute advantage as to the former and a relative advantage as to the later after 1949) and also in their ability to appeal to third parties (where the USSR had a potential advantage). Each had very different models about conflict. It is foolish to assume they would act symmetrically. Gavin Duffy, Seth Tucker, *Investigation of the Potential Contribution of AI Methods to the Avoidance of Crises and Wars,* Social Science Computing Review (Spring, 1995)

[15] Gary King, Brent Heeringa, David Westbrook, Joe Catalano, Paul Cohen, "Models of Defeat", Proceedings of the 2002 Winter Simulation Conference (2002) p. 928.

**B. Existing Research**

Existing research in artificial intelligence to represent political interactions seems more broad ranging than deep. That is, many different approaches have been tried but none appear to have been explored deeply. This is partly because the field of political interaction is itself very broad reaching: politics is universal and occurs at all times and places in the world's history. Topics consider range from the very abstract Ideology Machine[19] and policy arguing programs[20] to quite concrete scenarios of specific historical events.[21] More recently, data mining of political events has also captured the attention of researchers:[22] The Austrian

---

[16] *Id.* at 929.
[17] *Id.*
[18] *Id.* at 931.
[19] Gavin Duffy, Seth Tucker, *Investigation of the Potential Contribution of AI Methods to the Avoidance of Crises and Wars,* Social Science Computing Review (Spring, 1995)
[20] The Policy Arguer, Gavin Duffy, Seth Tucker, *Investigation of the Potential Contribution of AI Methods to the Avoidance of Crises and Wars,* Social Science Computing Review (Spring, 1995)
[21] E.g., the Cuban Missile Crisis, Gavin Duffy, Seth Tucker, *Investigation of the Potential Contribution of AI Methods to the Avoidance of Crises and Wars,* Social Science Computing Review (Spring, 1995).
[22] .
 Klaus Kovar, Johannes Fürnkranz, Johann Petrak, Bernhard Pfahringer, Robert Trappl, Gerhard Widmer
 Searching for Patters in Political Event Sequences: Experiments with the KEDS Database
 Cybernetics and Systems 31(6) 2000.

Research Institute for Artificial Intelligence (ARIAI)[23] is doing active research on the question of political AI, largely involving data mining. Only some of their research work is available on-line and the topic, data-mining, is outside the scope of this paper, thus the ARIAI is only mentioned as a possible point for scientific contact.

Existing research may be broad yet shallow because few social scientists are computer programmers. Thus, the existing base of programmers may have limited political expertise.[24] Finally, the problems of peace, while vital, are also not necessarily well funded since the results are less certain. Getting a government to fund a research project on robot tanks is much easier than funding a project with less concrete exciting and immediate results. Governments are more interested in funding war fighting than war prevention.

We can analyse existing research by considering the scenarios it addresses, the technologies it deploys, and existing implementations. This typology is suggested because existing surveys are inexhaustive[25] and have not yet developed a solid taxonomy.[26]

### 1. Scenarios

A wide variety of scenarios have been considered by political AI. Representative examples include the Cuban Missile Crisis[27] and the U.S. intervention in the Dominican Republic.[28]

---

[23] http://www.ai.univie.ac.at/oefai/aisoc/peace.html
[24] Mallery (1988).
[25] "We do not claim to present an exhaustive survey." Gavin Duffy, Seth Tucker, *Investigation of the Potential Contribution of AI Methods to the Avoidance of Crises and Wars,* Social Science Computing Review (Spring, 1995)
[26] "In reviewing the main AI applications in political science, we confess our inability to categorize these efforts neatly. We can provide only a loose categorization, founded more on family resemblances than any formal set of necessary and sufficient conditions for category membership." Gavin Duffy, Seth Tucker, *Investigation of the Potential Contribution of AI Methods to the Avoidance of Crises and Wars,* Social Science Computing Review (Spring, 1995)
[27] Gavin Duffy, Seth Tucker, *Investigation of the Potential Contribution of AI Methods to the Avoidance of Crises and Wars,* Social Science Computing Review (Spring, 1995).
[28] *Id*.

These scenarios serve to inform policy makers after the fact of their mistakes and as aides to human decision making may serve a useful role. There is really no limit to the scenarios that could be usefully modelled.

## 2. Technologies

The technologies used for implementation have generally focussed primarily on rule based expert systems. Secondarily, logic programming is also used, most often with PROLOG.[29] My research has not revealed scientific studies of multi-agent approaches, though CONSIM seems to be an agent based approach using a system of production. RISK and DIPLOMACY are however clear examples of multi-agent political games. Unfortunately, I have not found any scientific studies of these games.

## 3. Implementations
RISK

Risk is a multi-player (2-6) zero sum game with perfect information but stochastic determination of combat results. When several players are involved they will form coalitions to prevent the dominance of any one player. When a dominant player emerges several or even all other players will unite against that player weakening the dominant player to the point that some other dominant player emerges. The remaining players will then unite against the new dominant player. However since Risk is a zero sum game weaker players are ultimately eliminated leaving fewer and fewer players to form a coalition against the dominant player. This is because players in a coalition can defect from the coalition or even betray members of the coalition. This interplay of alliance, defection, and outright treachery does seem to roughly reflect international political (as opposed to economic) relations: the Napoleonic wars which Risk approximately models, featured a series of shifting coalitions ultimately resulting in the emergence of one dominant hegemon.

---

[29] *Id*.

Risk has been fairly succesfully represented using AI. The human player can outwit the AI but the AI is challenging enough to be interesting and the human player sometimes loses because there is a fair amount of luck involved in Risk.

### DIPLOMACY

Another game which represents international affairs using multiple agents is Diplomacy. Unlike Risk there are no random elements in diplomacy. Also unlike Risk information in Diplomacy is imperfect: all players moves are hidden from all other players and are executed simultaneously. Since each player must rely on other players to have sufficient power to advance their interests Diplomacy is a better game for the study of coalitions. Unfortunately, the AI in Diplomacy is severely limited: computer opponents are easily defeated by human players.[30]

### BALANCE OF POWER

Another interesting game simulating international politics is Balance of Power[31] In this game a player can implement a foreign policy to advance the interests of the U.S. or U.S.S.R. Implementation of an aggressive policy can result in escalation leading to nuclear war: the centerpiece of the game is the prevention or deescalation of nuclear conflicts. Yet at the same time the players must risk nuclear war to advance their interests. It is a game of brinksmanship. The computer AI allows a multiplayer mode where minor powers may enter into coalitions.

---

[30] Chris Harding "Diplomacy" Adrenaline Vault (Dec. 28 , 1999) http://www.avault.com/reviews/review_temp.asp?game=diplomacy

[31] Chris Crawford, Balance of Power: Microprose (1989).

## CONSIM

Whether computers can model political decision making was at first questioned[32] though over time that threshold question seems no longer to be asked. A computer can in all events model basic questions such as "what will country A do in reaction to policy 1 of country B?"[33] That is, each agent in our state system will have a policy - say wealth maximization, security maximization, ideological self replication, or some mix of these or other goals.

Central to the analysis of political decision making are risk, cost and benefit analysis.[34] Determination of estimated states depends on placing values (utiles) on states - even apparently subjective emotional states or on ambiguous states such as moral or religious values.[35] These intangibles however can be expressed because states do agree on basic principles such as economic prosperity, education and even reduction of violence.[36] Risk, cost and benefit are used to determine payoff of states in the system.

In CONSIM the authors constructe a matrix of probabilities. Each country has a "menu" of policies which it can choose from. It opponent may then, in reaction choose from its menu of policies.[37] Policies range from peaceful to aggressive including the option of nuclear war. This is very similar to the later BALANCE OF POWER but with a much smaller database and menu of actions. CONSIM examines all available information, determines possible reactions, evaluates the merit of those reactions and then chooses the best option available based on its goals.[38] That seems much like the behavior of an agent.

---

[32] Joe Clema, John Kirkham, *CONSIM (Conflict Simulator): Risk, Cost and Benefit in Political Situations,* p. 226.
[33] *Id.* at 227.
[34] *Id.* at 227.
[35] *Id.*
[36] *Id.*
[37] *Id.* at 228.

The authors of CONSIM believe that future political simulations will essentially extend these basic characteristics.[39] The authors predict descendants of CONSIM will be better than humans in making decisions because all available information can be evaluated.[40] However that is not the case; a computer can evaluate more information than a human but all information cannot be considered due to the information revolution sparked by computers. The authors also argue the computer will make better decisions because computers make no mistakes.[41] Again, this very optimistic view ignores the fact that even if a computer has perfectly running hardware it is only as accurate as the software written for it. The authors also believe increased computational speed will improve computer performance such that it will surpass humans.[42] Mallery disagrees and shows cogently that increased speed of computers can increase speed of errors made by computers: A small indetectable error in one operation iterated millions of times results in a huge error whose source is difficult to discover. The final reason offered for better AI performance is that prejudice and emotion are removed from decision making resulting in more objective outcomes.[43] That is partly correct: Computer analysis can help us to step back from the problem and be more realistic about it. However prejudice still exists and is programmed into the computer at the point where values are chosen and weights assigned to risks, costs and benefits. But that is not an argument that the computer can be a superior decision maker than a human. Rather it is an argument that the computer can help humans to reevaluate human decisions.

---

[38] *Id.* at 229.
[39] *Id.* at 230.
[40] *Id.* at 234.
[41] *Id.*
[42] *Id.*
[43] Id at 234.

**Critique**

CONSIM can be criticized as it is an irresponsible and frightful experiment which proposes that computers should be used to determine whether to use nuclear weapons.[44] Mallery rightly critiques such folly. He points out that even if the AI works, computerization leads to a risk of amplification of erroneous messages[45] For an example of this we can look at markets where the second order effects of AI have clearly resulted in increased volatility.[46] Mallery concludes that AI modeling is at present only suitable for description and research but is not adequate for prescriptive policy making.[47] I agree with this analysis: AI has an interesting pedagogical and even analytical role but should not replace a human decision maker in questions of war fighting.

## III. IMPLEMENTATION

The author proposes a simple three agent system to model very basic diplomatic and economic strategies. It consists of three "states", one which centers its strategy on national security, one centering its strategy on trade, and a third which takes a mixed approach of trade and security. The agents can build arms, enter into simple alliances, pay tribute to stronger states, or attack and predate the resources of weaker states. Utility functions are determined by the economic well being of the state: desperate agents do desperate things.

---

[44] John Mallery, Thinking about Foreign Policy: Finding an Appropriate Role for Artificially Intelligent Computers, 1998 Annual Meeting of the International Studies Association (1988)

[45] "As the time required to take actions and react decreases, the rate at which actions and reactions can occur increases. This increases the gain in the system which in turn, increases the probability of non-linear amplification of small intiial perturbations in strategic systems. Thus, even if the AI system works correctly, the presence of these systems can increase gain, and therefore, lower the stability of international security sytems." *Id*.

[46] *Id*.

[47] *Id.*

Internationally, states rarely disappear completely. Poland is just one example of a state which has often been dismembered only to be reconstructed. Thus, in the game, even if a state in this game loses all its resources or arms it continues to exist. Thus in a sense this is not a "game" since we usually think that games have an end state. Thus we would choose an arbitrary numebr of iterations for the "end state": Say, ten turns.

This raises the question how is victory determined? We know that a state with fewer resources and weapons than its neighbor has lost, but what of two states, one of which is better armed but poorer than the other? An experimental approach would run the simulation one turn further. That is in fact one point the game is trying to illustrate: the indeterminate relations of states. In absolute terms, both the mercantile state Athens and the militarist state Sparta were worse after the Pelopenesian war than before. Yet in relative terms, Sparta emerged militarily dominant but culturally poor. Everyone lost by any measure in absolute terms but in relative terms of power (as opposed to wealth) Sparta won.

**A. Agent Strategies**

The strategies of the three agents are very simple: the mercantile agent seeks to maximize its wealth building arms only when absolutely necessary. The mercantile state can also seek to appease its more aggressive neighbors and always seeks to trade with them rather than predate their resources. It is a "sharper" trader in that it will always outperform its rivals in trading in relative terms. That is, trade in the game is positive sum in absolute terms, however it is more beneficial to the mercantile state than the other states.

The militarist state in contrast seeks to achieve military dominance over its opponents. It can and will attempt to predate the resources of its neighbors at every opportunity using the stolen resources to build more arms to steal more resources. The militarist state does not seek to enter into alliances becuase it is following a strategy of autarchy.

The mixed state attempts to both trade and maintain a strong military. It can predate the resources of weaker states but prefers to trade because it can build more arms by trading resources and then converting traded resources than by simply predating. The mixed state does not pay tribute to the mercantile state but can pay tribute to the militarist state.

**B. Agent Intentions**

No state can observe the intentions of any other state, however the game presumes perfect information: each state knows the size of the armies and economy of the other states.

**C. Learning Functions**

The agents in this game can learn: namely, if a state attempts to pay tribute, but is predated anyway it will no longer pay tribute. A more developed game would have the states remember what states entered into and observed alliances, and would allow states having formed alliances to break them treacherously or openly.

**D. Results**

In principle, the game presents a simplified model of inter-state relations. It could be made more realistic through the addition of further agents (states), a closer linkage between utiles used in the game and actual economic and military statistics. Formation and renunciation of alliances could also be integrated into the game. But at least the rudiments of a model of international relations are represented by this game. The game illustrates some useful and interesting features both of international relations and of multi-agent approaches to AI.

**E. Paths for future research**

Future AI representations of international politics should examine the multi-agent approach sketched here in greater detail. Existing implementations should be studied and expanded upon to consider the fact that international relations are at times zero sum but generally are in fact positive sum. It may be possible to use open source implementations of various Risk like games as a basis for more serious research on shifting alliances in coalition game theory.

## **CONCLUSIONS**

Though political AI cannot be used for war fighting it can be used as an aid to human decision making, as an evaluative heuristic for planing and review. Moreover, political AI has pedagogic interest. Simulations of international relations have been proven to create more interest in students than conventional lecturing.[48] Further, presenting international relations problems as simulations results in students having a greater understanding of problems in international relations and how decisions are made.[49] Simulations may reduce the learning time of students in international politics.[50] What lessons does political AI teach?

---

[48] David Louscher, The Effectiveness of a Short Term Simulation for Teaching Foreign Policy and National Security Affairs, Simulation and Games (December 1977) p. 449.
[49] *Id.*
[50] *Id.*

We have seen the errors in planning (failing to plan for war termination; presuming symmetry of opponents where it may not in fact exist) and the risks of AI in game playing (automated nuclear war). Simulation results also reveal something else which is just as interesting: Zero sum games representing international affairs tend "to produce universal empires, in which one state eventually achieves global domination."[51] Unlike the discredited balance of power theories, which argued that shifting coalitions would prevent the emergence of one universal power, universal empires seem to be the fate of the world: Rome, but also China and the Incas and Aztecs even the Moguls show us historical examples of universal empire. More recently the British empire and most currently American hegemony show the tendancy of the world to a monopoly of power. AI reveals what is already implicit in a historical understanding of international relations. However it does not tell us explicitly why multi-polar systems tend to monopoly of power under one hegemon nor whether that is economically desirable. Finally, implementations are zero sum and thus one dimensional, focussing almost exclusively on political power (the ability to deploy violent force) without considering economic power.

My implementation is intended to illustrate:

1) a multi-agent approach

2) a mixed game approach:

a) zero sum aspects as to deterrence

b) negative sum aspects as to war fighting

c) positive sum aspects as to trade

I hope that this account will result in a more balanced understanding of how states interact with each other than the bellophilic examples. In this regard, we should look carefully at

---

[51] Gavin Duffy, Seth Tucker, *Investigation of the Potential Contribution of AI Methods to the Avoidance of Crises and Wars,* Social Science Computing Review (Spring, 1995).

Crawford's Balance of Power - a zero sum wargame where the goal of the game is to prevent war.

# BIBLIOGRAPHY

AIAIR http://www.ai.univie.ac.at/oefai/aisoc/peace.html

Joe Clema, John Kirkham, *CONSIM (Conflict Simulator): Risk, Cost and Benefit in Political Situations,*

Chris Crawford, Balance of Power: Microprose (1989).

Gavin Duffy, Seth Tucker, *Investigation of the Potential Contribution of AI Methods to the Avoidance of Crises and Wars,* Social Science Computing Review (Spring, 1995)

Gavin Duffy, Seth Tucker, The Policy Arguer, *Investigation of the Potential Contribution of AI Methods to the Avoidance of Crises and Wars,* Social Science Computing Review (Spring, 1995)

Andrew Grosso, *The Demise of Sovereignty*, 44/3 Communications of the ACM (2001) p. 102.

Chris Harding "Diplomacy" Adrenaline Vault (Dec. 28 , 1999)
http://www.avault.com/reviews/review_temp.asp?game=diplomacy

Thomas Hobbes, Leviathan, ch. 14 (1660) http://oregonstate.edu/instruct/phl302/texts/hobbes/leviathan-c.html#CHAPTERXIV

Klaus Kovar, Johannes Fürnkranz, Johann Petrak, Bernhard Pfahringer, Robert Trappl, Gerhard Widmer
Searching for Patters in Political Event Sequences: Experiments with the KEDS Database
Cybernetics and Systems 31(6) 2000.

Hume, David *Essays, Moral, Political, and Literary* Part II, Essay VII, "Of the Balance of Power" (1742)
http://www.econlib.org/library/LFBooks/Hume/hmMPL30.html#Part%20II,%20Essay%20VII,%20OF%20THE%20BALANCE%20OF%20POWER

Gary King, Brent Heeringa, David Westbrook, Joe Catalano, Paul Cohen, "Models of Defeat", Proceedings of the 2002 Winter Simulation Conference (2002) p. 928.

David Louscher, The Effectiveness of a Short Term Simulation for Teaching Foreign Policy and National Security Affairs, Simulation and Games (December 1977) p. 449.

John Mallery, "Thinking about Foreign Policy: Finding an Appropriate Role for Artificially Intelligent Computers", 1998 Annual Meeting of the International Studies Association (1988)

Christos Papadimitriou, *Algorithms, Games and the Internet,* STOC'01 (2001), p. 749-750  2001 ACM 1-58113-349-9/01/0007.

David Ricardo, *On The Principles of Political Economy and Taxation*, Ch. 7 (1817)
http://www.marxists.org/reference/subject/economics/ricardo/tax/ch07.htm

Ted Scully, Micheal Maden, Gerard Lyons "Coalition Calculation in a Dynamic Agent Environment" Proceedings of the 21st International Conference on Machine Learning, Banff, Canada (2004).

Adam Smith, *On the Nature and Causes of the Wealth of Nations* (1776)
http://www.econlib.org/library/Smith/smWN.html

Wikipedia "Emergence" (2004) http://en.wikipedia.org/wiki/Emergence

Wikipedia "Balance of Power" (2004) http://en.wikipedia.org/wiki/Balance_of_power